\RequirePackage{setspace}
\documentclass[sigconf]{acmart}
\usepackage{stfloats}
\usepackage{multirow}
\usepackage{subfigure} 
\usepackage{makecell}
\usepackage{caption}
\usepackage{subfigure}
\usepackage{arydshln}
\usepackage{booktabs}
\usepackage{multirow}
\usepackage{tablefootnote}
\usepackage{setspace}
\usepackage{hyperref}
\usepackage{amsmath}
\usepackage{ulem}
\usepackage{multirow}
\usepackage{arydshln}
\usepackage{booktabs}
\usepackage{newfloat}
\usepackage{listings}
\usepackage{bm}
\usepackage[utf8]{inputenc} % allow utf-8 input
\usepackage[T1]{fontenc}    % use 8-bit T1 fonts
\usepackage{hyperref}       % hyperlinks
\usepackage{url}            % simple URL typesetting
\usepackage{booktabs}       % professional-quality tables
\usepackage{amsfonts}       % blackboard math symbols
\usepackage{nicefrac}       % compact symbols for 1/2, etc.
\usepackage{microtype}      % microtypography
\usepackage{xcolor}         % colors
\usepackage{graphicx}
\usepackage{subfigure}
\usepackage{amsmath}
\usepackage{color}
\usepackage{soul} 
\usepackage{xcolor}
\usepackage{subfigure}
\usepackage[ruled,vlined]{algorithm2e}
\usepackage{wrapfig}

\usepackage{arydshln}
\usepackage{booktabs}
\usepackage{multirow}
\definecolor{hl}{RGB}{173,216,230}
\AtBeginDocument{%
  \providecommand\BibTeX{{%
    \normalfont B\kern-0.5em{\scshape i\kern-0.25em b}\kern-0.8em\TeX}}}

\copyrightyear{2024} 
\acmYear{2024} 
\setcopyright{rightsretained} 

\acmConference[WWW '24]{Proceedings of the ACM Web Conference 2024}{May
13--17, 2024}{Singapore, Singapore}
\acmBooktitle{Proceedings of the ACM Web Conference 2024 (WWW '24), May
13--17, 2024, Singapore, Singapore}\acmDOI{10.1145/3589334.3645363}
\acmISBN{979-8-4007-0171-9/24/05}

\usepackage{etoolbox}
\makeatletter
\patchcmd{\maketitle}{\@copyrightpermission}{
  \begin{minipage}{0.3\columnwidth}
     \href{https://creativecommons.org/licenses/by/4.0/}{\includegraphics[width=0.90\textwidth]{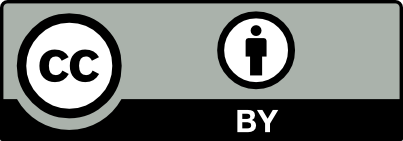}}
  \end{minipage}\hfill
  \begin{minipage}{0.7\columnwidth}
     \href{https://creativecommons.org/licenses/by/4.0/}{This work is licensed under a Creative Commons Attribution International 4.0 License.}
  \end{minipage}
  
  \vspace{5pt}
}{}{}

\makeatother

\begin{document}

\title{Search-in-the-Chain: Interactively Enhancing Large Language Models with Search for Knowledge-intensive Tasks}

\author{Shicheng Xu}
\orcid{0000-0001-7157-3410}
\affiliation{%
   \institution{CAS Key Laboratory of AI Security, \\ Institute of Computing Technology, Chinese Academy of Sciences \\ University of Chinese Academy of Sciences}
  %\country{University of Chinese Academy of Sciences, Beijing, China}  }
  \city{Beijing}
  \country{China}  }
\email{xushicheng21s@ict.ac.cn}

\author{Liang Pang}
\authornote{Corresponding author}
\affiliation{%
   \institution{CAS Key Laboratory of AI Security, \\ Institute of Computing Technology, Chinese Academy of Sciences}
  \city{Beijing}
  \country{China}
  }
\email{pangliang@ict.ac.cn}

\author{Huawei Shen}
\affiliation{%
   \institution{CAS Key Laboratory of AI Security, \\ Institute of Computing Technology,}
  \country{Chinese Academy of Sciences \\ Beijing, China}}
\email{shenhuawei@ict.ac.cn}

\author{Xueqi Cheng}
\affiliation{%
  \institution{CAS Key Laboratory of AI Security, Institute of Computing Technology, Chinese Academy of Sciences}
  \country{Beijing, China}}
\email{cxq@ict.ac.cn}

\author{Tat-Seng Chua}
\affiliation{%
   \institution{Sea-NExT Joint Lab, National University of Singapore}
  \country{Singapore}}
\email{dcscts@nus.edu.sg}

\begin{abstract}
Making the content generated by Large Language Model (LLM), accurate, credible and traceable is crucial, especially in complex knowledge-intensive tasks that require multi-step reasoning and each step needs knowledge to solve. Retrieval-augmented generation is good potential to solve this problem. However, where and how to introduce Information Retrieval (IR) to LLM is a big challenge. Previous work has the problems that wrong knowledge retrieved by IR misleads the LLM and interaction between IR and LLM breaks the reasoning chain of LLM. This paper proposes a novel framework named \textbf{Search-in-the-Chain} (SearChain) for the interaction between LLM and IR to solve the challenges. First, LLM generates the reasoning chain named Chain-of-Query (CoQ) where each node consists of an IR-oriented query-answer pair. Second, IR verifies the answer of each node of CoQ. It corrects the answer that is not consistent with the retrieved information when IR gives high confidence, which improves the credibility. Third, LLM can indicate its missing knowledge in CoQ and rely on IR to provide this knowledge to LLM. These operations improve the accuracy in terms of reasoning and knowledge. Finally, SearChain generates the reasoning process and marks references to supporting documents for each reasoning step, which improves traceability. Interaction with IR in SearChain forms a novel reasoning path based on a tree, which enables LLM to dynamically modify the direction of reasoning. Experiments show that SearChain outperforms state-of-the-art baselines on complex knowledge-intensive tasks including multi-hop Q\&A, slot filling, fact checking, and long-form Q\&A.
\end{abstract}



%%
%% Keywords. The author(s) should pick words that accurately describe
%% the work being presented. Separate the keywords with commas.

\maketitle

\normalem
\section{Introduction}
\label{intro}
Large Language Models (LLMs) such as ChatGPT have shown promising performance in various natural language processing tasks~\cite{test_llm1,surveyllm}. However, for the complex knowledge-intensive tasks that require multi-step reasoning and each step needs knowledge to solve~\cite{kilt,kltask,zhu2021adaptive}, many studies have shown that LLMs have trouble in: (1) compositional reasoning over multiple knowledge~\cite{selfask}, (2) memorization of long-tail and real-time knowledge~\cite{long-tail} and (3) avoiding hallucination that is inconsistent with the facts~\cite{huanjue}, which affects the accuracy and credibility of LLMs for complex knowledge-intensive tasks. Besides, context-only generation without any supporting evidence causes less traceability and makes people less trust in the LLM-generated content. Retrieval-augmented method has good potential to solve these problems because it combines the knowledge of the model with external knowledge bases~\cite{fid,rag,realm}. 

However, where and how to introduce IR into LLM is not a trivial thing. There are three main challenges. \textcolor{red}{\bm{$\mathcal{C}\mbox{-}1$}}: Directly inserting IR into the reasoning process of LLM such as Self-Ask~\cite{selfask}, LTM~\cite{leasttomost}, React~\cite{react} and DSP~\cite{dsp} leads to breaking the reasoning chain of LLM. Because in these methods, LLM can only reason a local sub-question in each generation. \textcolor{red}{\bm{$\mathcal{C}\mbox{-}2$}}: When there is a conflict in the knowledge of IR and LLM, for the knowledge that the LLM has correctly memorized, it risks being misled by IR if IR retrieves the wrong information. It is important to make sure that IR only provides the knowledge that LLM really needs. \textcolor{red}{\bm{$\mathcal{C}\mbox{-}3$}}: Previous methods cannot dynamically modify the reasoning direction.

\begin{figure*}[t]
\centering
\includegraphics[width=\linewidth]{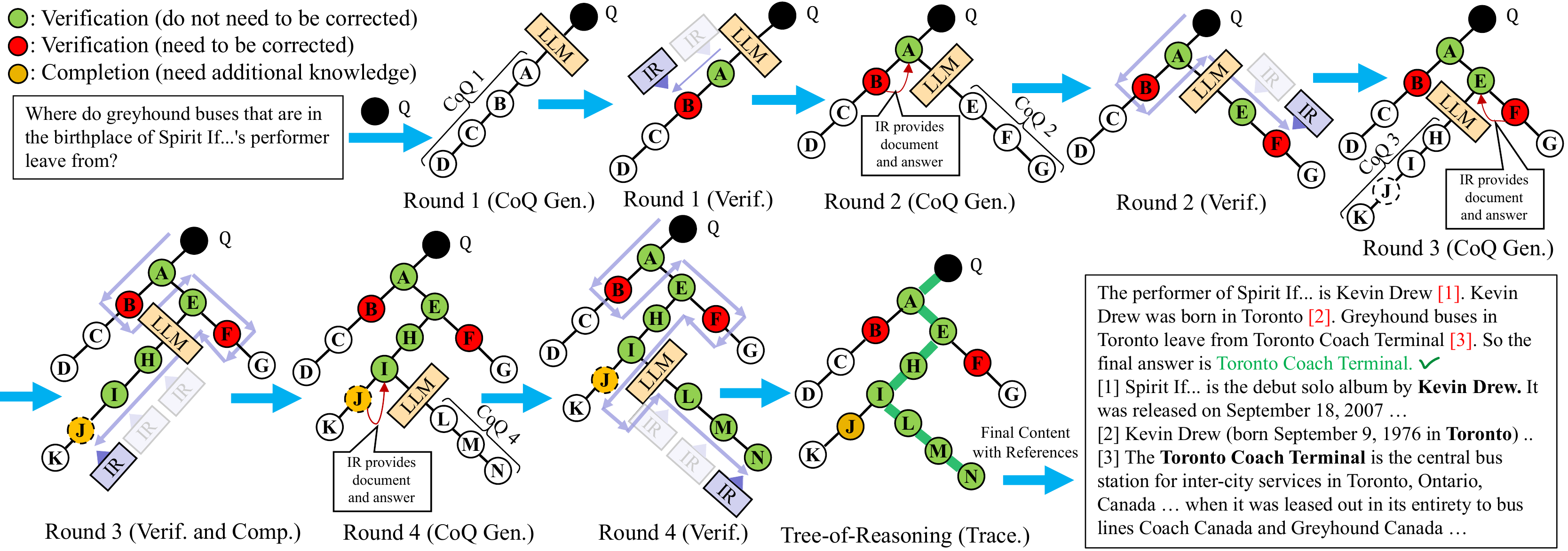} % Reduce the figure size so that it is slightly narrower than the column.
\caption{Interaction between IR and LLM in SearChain. First, SearChain makes LLM plan a CoQ where each node is a query-answer pair. Then, IR interacts with each node of CoQ to perform verification and completion. If IR detects that a node needs to be corrected or provided with knowledge, it gives feedback to LLM and LLM re-generates a new CoQ, which is the new branch of the tree. This process is the node-identify Depth-first Search on a tree called Tree-of-Reasoning (the correct reasoning path is green). The final content includes the reasoning process and references to supporting documents.}
\label{motivation}
\end{figure*}

In this paper, we propose a novel framework named Search-in-the-Chain (SearChain) to effectively combine LLM with IR to solve the above challenges (Figure~\ref{motivation}). SearChain and previous methods both need multiple IR-LLM interaction rounds, but the former works at the chain level, while the latter only deals with a node. In each round, SearChain performs reasoning, verification, and completion. After the interaction, SearChain performs tracing to generate the final content. Specifically, in each round, first, LLM exploits in-context learning to construct a Chain-of-Query (CoQ), which is a reasoning chain to decompose and solve complex questions. Each node of the chain consists of an IR-oriented query, the answer generated by LLM for this query, and a flag indicating whether LLM needs additional knowledge. Different from previous methods in which LLM can only perform one-step reasoning (only a node) when interacting with IR, CoQ is a complete chain. This design avoids IR from breaking the reasoning chain (\textcolor{red}{\bm{$\mathcal{C}\mbox{-}1$}}). Second, IR interacts with each node of CoQ to perform verification and completion. In verification, IR verifies the answer of each node. In case when the LLM-generated answer is not consistent with the retrieved information and IR gives high confidence, IR gives feedback to LLM to help it correct the answer and re-generate the correct CoQ. In completion, IR determines whether the node has missing knowledge from the flag of the node and provides this knowledge to LLM to help it re-generate CoQ. LLM gradually generates the correct CoQ through multiple rounds of interaction with IR. The above design provides LLM with the knowledge it really needs to alleviate the misleading caused by IR to LLM (\textcolor{red}{\bm{$\mathcal{C}\mbox{-}2$}}), which improves accuracy. IR verifies and corrects the knowledge in the reasoning process of LLM based on external knowledge bases, which improves credibility. After the interaction, SearChain performs tracing to generate the reasoning process and marks references to supporting documents for each reasoning step, which is used as the final content returned to the user. This improves the traceability of knowledge in the generated content. Interaction with IR in SearChain transforms the reasoning path from a chain to node-identify Depth-first Search on a tree called Tree-of-Reasoning (ToR). CoQ generation can be seen as a part of Depth-first Search and IR can identify the nodes that need more information (\textcolor{red}{\bm{$\mathcal{C}\mbox{-}3$}}). This enables LLM to dynamically modify the reasoning direction. This paper's main contributions are:

\textbf{(1)} We highlight the challenges in introducing IR into LLM from the perspectives of reasoning and knowledge. 

\textbf{(2)} SearChain not only improves the knowledge-reasoning ability of LLM but also uses IR to identify and give the knowledge that LLM really needs. Besides, SearChain can mark references to supporting documents for the knowledge involved in the generated content. 

\textbf{(3)} Interaction with IR in SearChain forms a novel reasoning path: node-identify Depth-first Search on a tree, which enables LLM to dynamically modify the direction of
reasoning.

\textbf{(4)} Experiment shows that SearChain outperforms state-of-the-art baselines on complex knowledge-intensive tasks including multi-hop Q\&A, slot filling, fact checking and long-form Q\&A. Code is released at \url{https://github.com/xsc1234/Search-in-the-Chain}.

\section{Related Work}
% In this section, we discuss related work on Chain-of-Thought Prompting and Retrieval-augmented Language Models.
\subsection{Chain-of-Thought Prompting} Chain-of-thought~\cite{chain} proposes the method that uses few-shot examples to enable LLM to give intermediate reasoning results to improve the reasoning ability.~\cite{chain4} uses "Let's do it step by step" as prompt to achieve promising zero-shot performance. Auto-CoT exploits language models to automatically construct few-shot learning examples for CoT~\cite{chain5}. There are also many studies that cover other aspects of CoT such as self-consistency~\cite{chain2}, usage of small and medium size models~\cite{chain3} and  selection~\cite{complex}. Besides, there are studies that iteratively use LLM to decompose complex questions and answer sub-questions step by step. These methods include Least-to-Most~\cite{leasttomost}, Dynamic Least-to-Most~\cite{dynamic}, Self-Ask~\cite{selfask} and DSP~\cite{dsp}. Chain-of-Query of our method is also inspired by CoT. However, previous studies focus on giving intermediate reasoning results or decomposing complex questions and answering sub-questions step by step. They focus on how to solve local sub-questions while ignoring the global planning of the reasoning chain. Although AgentGPT and PS~\cite{ps} first plan each sub-question and then solve them, they are not suitable for scenarios where the next sub-question needs the answer of the previous sub-questions to generate, which is common for complex knowledge-intensive tasks (multi-hop QA). CoQ of our method makes LLM construct a global reasoning chain where each node is a query-answer pair. This design not only improves the knowledge-reasoning ability but also provides the interface for IR to be deeply involved in the reasoning process of LLM. 

\subsection{Retrieval-augmented Language Models} Many studies have shown that retrieval-augmented methods get promising performance in various natural language tasks such as open-domain question answering~\cite{fid,rag,realm,mou2021narrative,cheng2010uncovering,xu2022match,xu2024list}, language modeling~\cite{nmlm,retro,niu2012top} and enhancing the factuality~\cite{webbrain}. Recently, some studies enable LLM to interact with IR via in-context learning~\cite{selfask,dsp,react,toolformer}. In these methods, the interaction between IR and LLM makes the reasoning of LLM not continuous. LLM can only perform one-step reasoning at each inference. Our method makes LLM generate a global reasoning chain called Chain-of-Query at each inference, which introduces stronger logical relationship between each reasoning step. Besides, previous methods can only provide information to the LLM but cannot assist LLM in correcting erroneous information or avoid the negative effect of IR on LLM, which makes the reasoning of LLM still in a one-dimensional chain. Our method makes IR interact with each node of the chain. IR only provides LLM with its missing knowledge and corrects the answers that are not consistent with the retrieved information when IR is confident enough. This mitigates the negative effect of IR on LLM and transforms the reasoning path from chain to node-identify Depth
First Search on a tree to enable LLM to dynamically modify the reasoning direction.

\section{Our Method}
\label{method}
This section introduces the design of Search-in-the-Chain (SearChain). In SearChain, IR and LLM conduct multiple rounds of interaction. In each round, first, LLM acts as the commander to plan the global reasoning chain for the complex input questions called Chain-of-Query (CoQ). Each node of the CoQ consists of an IR-oriented query, the answer for this query, and a flag indicating whether LLM needs additional knowledge. Then, IR interacts with each node of CoQ and performs the completion and verification to only provide LLM with missing knowledge and correct the wrong answers to alleviate the misleading. LLM re-generates new CoQ based on feedback from IR. Multiple rounds of interaction help LLM gradually generate the correct CoQ, which improves accuracy and credibility. Finally, SearChain performs tracing to generate the whole reasoning process and marks references to supporting documents for each reasoning step, which is used as the final content returned to the user. This improves the traceability of generated content. Interaction with IR in SearChain transforms the reasoning path from a chain to node-identify Depth-first Search on a tree called Tree-of-Reasoning (ToR), which enables LLM to dynamically modify the reasoning direction.  

% LLM with the knowledge it really needs to alleviate the misleading of LLM. 

\subsection{Comparison with Previous Methods}
\begin{figure}[t]
\centering
\includegraphics[width=\linewidth]{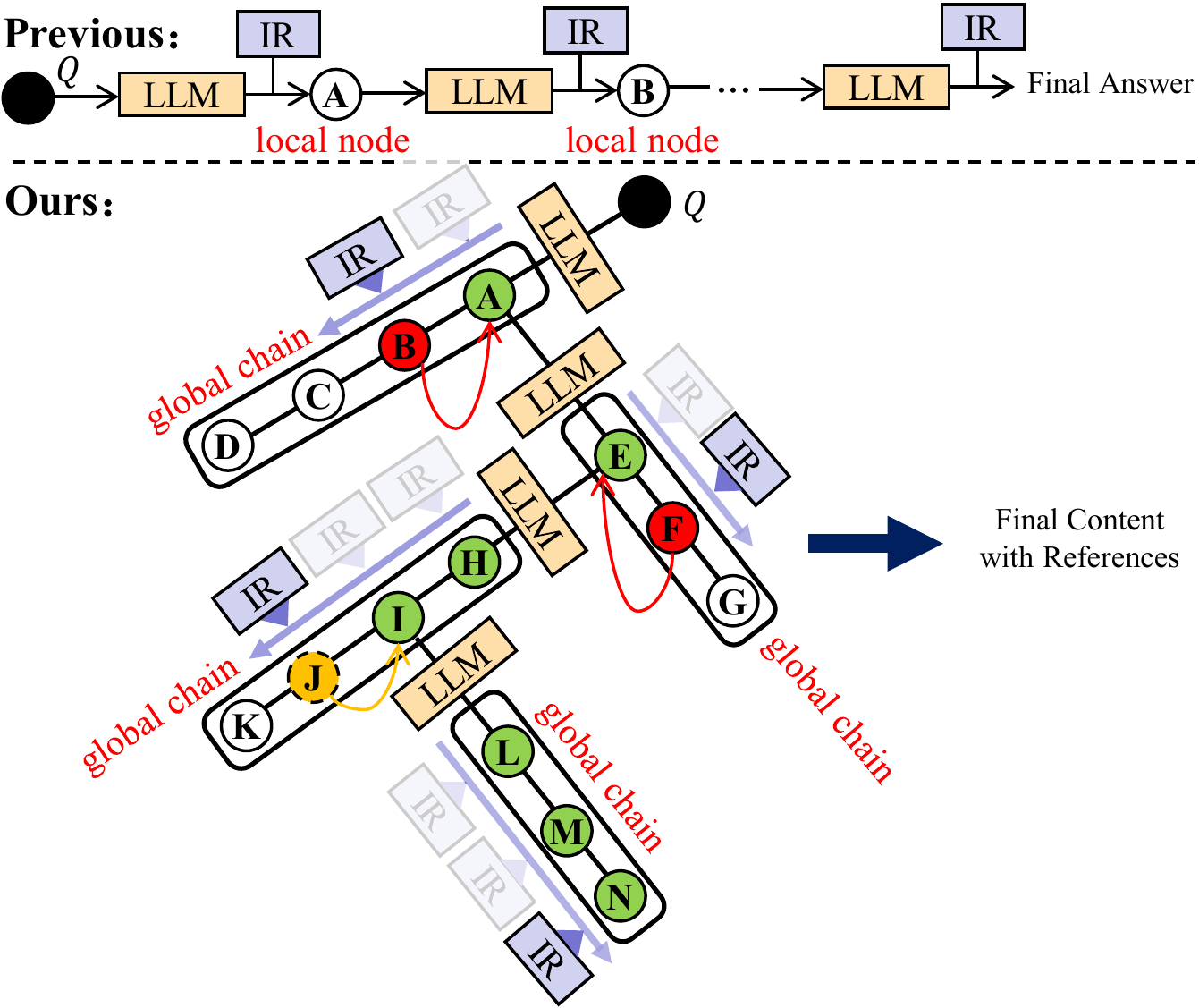} % Reduce the figure size so that it is slightly narrower than the column.
\caption{Comparison with previous methods.}
\label{comp}
\end{figure}
Figure~\ref{comp} shows the difference between our method and previous retrieval-augmented methods (Self-Ask~\cite{selfask}, React~\cite{react}, DSP~\cite{dsp}, etc.) in solving complex knowledge-intensive questions. 

\noindent \textbf{(1) Local vs. Global.} For a complex question that needs multi-step reasoning, previous methods directly insert IR into the multi-step reasoning process, causing LLM can only reason a \textbf{local} sub-question such as node \textcircled{A} in each generation. This breaks the reasoning chain of LLM. Our method proposes Chain-of-Query to provide the interactive interface for IR on the premise of ensuring the coherence of reasoning chain (plans a \textbf{global} chain for question $Q$ such as \textcircled{\raisebox{-0.2pt}{A}}$\rightarrow$\textcircled{\raisebox{-0.8pt}{B}}$\rightarrow$\textcircled{\raisebox{-0.8pt}{C}}$\rightarrow$\textcircled{\raisebox{-0.7pt}{D}} in each generation). (solves \textcolor{red}{\bm{$\mathcal{C}\mbox{-}1$}})

\noindent \textbf{(2) Directly Provide vs. Verify and Complete.} Previous methods directly provide the retrieved information to the LLM. When the retrieved information is incorrect, the LLM runs the risk of being misled. In our method, IR only corrects inconsistent information in Chain-of-Query when IR is confident enough, and provides the information that LLM does not know via flags on Chain-of-Query, which mitigates the negative effect of IR on LLM. (solves \textcolor{red}{\bm{$\mathcal{C}\mbox{-}2$}})

\noindent \textbf{(3) Chain vs. Tree.} Previous methods cannot modify the reasoning direction in time as necessary. Our method transforms the reasoning path from a chain to node-identify Depth-first Search on a tree by introducing the verification and completion from IR, which enables LLM to dynamically modify the direction of reasoning. (solves \textcolor{red}{\bm{$\mathcal{C}\mbox{-}3$}})

\begin{figure}[t]
\centering
\includegraphics[width=\linewidth]{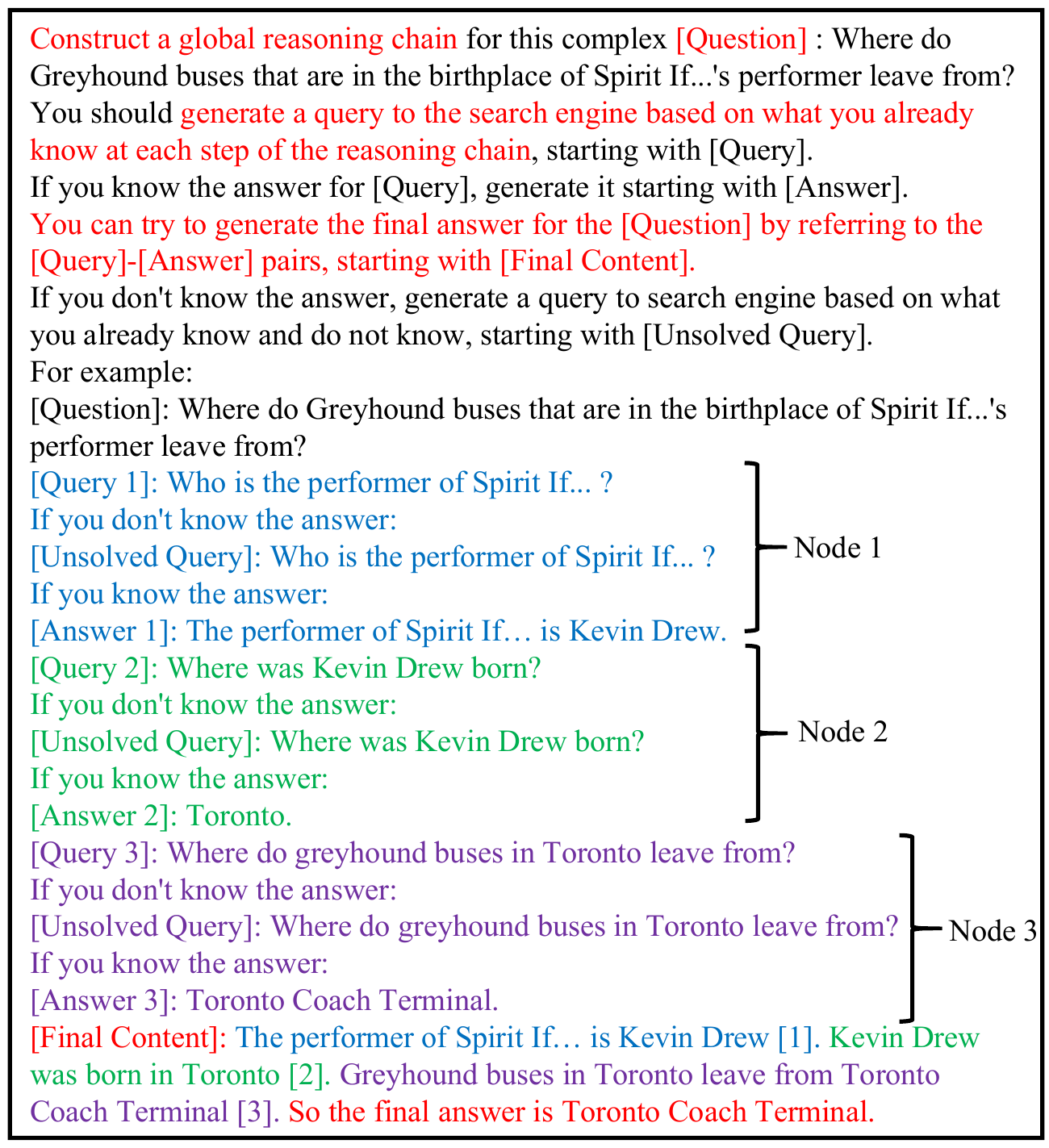} % Reduce the figure size so that it is slightly narrower than the column.
\caption{Prompt to make LLM generate Chain-of-Query.}
\label{prompt}
\end{figure}
\subsection{Chain-of-Query Generation}
\label{CoQ}
In SearChain, we use in-context learning~\cite{chain} to prompt large language model to construct a global reasoning chain for complex question $Q$ named Chain-of-Query (CoQ): 
\begin{align}
\textrm{CoQ} = (q_1,a_1)	\rightarrow (q_2,a_2) \rightarrow ... \rightarrow (q_n,a_n),
\end{align}
which is the branch of Tree-of-Reasoning. Each node $(q_i,a_i)$ of CoQ consists of an IR-oriented query $q_i$ and its answer $a_i$. $q_1$ ... $q_n$ are the sub-questions that need to be solved in the reasoning process of solving $Q$. CoQ generation is applied to each round of interaction between LLM and IR. In the first round, the prompt used to make LLM generate CoQ is shown in Figure~\ref{prompt}. The prompt with ``\textit{Construct a global reasoning chain}'' makes LLM know that the main task is to generate a global reasoning chain in each generation. \textbf{"Global"} means that LLM needs to plan a complete reasoning chain for the complex question, rather than answer the question directly or only solve \textbf{"local"} sub-questions (comparison shown in Figure~\ref{comp}). At each node of the chain, LLM focuses on generating the IR-oriented query and gives the answer if LLM knows. If LLM does not know the answer, it should mark the query with ``[Unsolved Query]'', which is a flag indicating the missing of knowledge. In subsequent rounds, when a node needs IR to correct or provide missing knowledge, LLM generates a new CoQ according to the feedback of IR to dynamically modify the reasoning direction. The design for this scenario will be introduced in Section~\ref{interaction}. The generation of CoQ is a complete Depth-first Search for $Q$, which avoids IR from breaking the reasoning chain of LLM. Experiments (Section~\ref{reasoning}) also show that for the difficult sub-question, CoQ enables LLM to solve it by more reasoning steps such as rewriting or further decomposing the sub-question while baselines tend to stop reasoning. It is because baselines focus on solving current local sub-questions while ignoring the global planning of the reasoning chain. The global perspective in CoQ makes LLM try more to explore possible answers when facing intermediate difficulties.

\begin{algorithm}[t]
	\caption{Description of the Interaction with IR.}
	\label{alg:algorithm1}
	\SetKwInOut{Initialize}{Initialize}
	\SetKwInOut{Output}{Output}
	\SetKwFunction{NotEqual}{NotEqual}
	\SetKwFunction{add}{add}
	\SetKwFunction{TemplateVerify}{TemplateVerify}
	\SetKwFunction{TemplateComlete}{TemplateComlete}
	\SetKwFunction{ChainGenerate}{ChainGenerate}
	\SetKwFunction{Tracing}{Tracing}
	\SetKwFunction{Retrieval}{Retrieval}
	\SetKwFunction{AddChild}{AddChild}
% 	\Input{Chain-of-Query: ${\boldsymbol{C}}$, Queries that has been solved: ${M^}$, Supporting documents: $R$}
	\Initialize{Processed queries: $M = null$;\\ Correct reasoning path: $R = null$;\\ Interaction rounds: $r = 0$;\\ Feedback: $ F= null$; ToR: $\boldsymbol{T} = Q$;}
	%\Output{Final Generated content.}  
	
    \SetKwFunction{DuplicateQuery}{DuplicateQuery}

    \SetKwFunction{PromptForVerify}{PromptForVerify}
    \SetKwFunction{PromptForComplete}{PromptForComplete}
    
	\SetKwFunction{IR}{IR}
    \SetKwProg{Fn}{Function}{:}{}
    \Fn{\IR{$q_i$, $a_i$}}{
    $d_i =$ \Retrieval{$q_i$};
    \textit{// Retrieve Top-1 document $d_i$ for $q_i$.}\\
    $g$, $f =$ Reader($q_i$, $d_i$)\;
    \textit{// Extract answer $g$ from $d_i$ and give confidence $f$.}\\
    \If{$q_i$ is Unsolved Query}
    {
    \textit{// Completion.} \\
    $R$.\add($q_i$, $g$, $d_i$);\quad \textit{// Record the correct node.}\\
    \textbf{return} \PromptForComplete{$q_i$, $g$, $d_i$}; %\tcp{Construct a prompt for completion.}
    }
    \If{$f > \theta$ and \NotEqual{$g, a_i$}} 
    {   \textit{// Verification.} \\
        $R$.add($q_i$, $g$, $d_i$);\quad \textit{// Record the correct node.}\\
        \textbf{return} \PromptForVerify{$q_i$, $g$, $d_i$}; %\tcp*{Construct a prompt to correct LLM.}
    }
    $R$.\add($q_i$, $a_i$, $d_i$); \textbf{return} ``Pass'' \;
    }
    
    \SetKwFunction{Traverse}{Traverse}
    \SetKwProg{Fn}{Function}{:}{}
    \Fn{\Traverse{$\boldsymbol{CoQ}$}}{
	\ForEach{$(q_i, a_i)$ in $\boldsymbol{CoQ}$}{
	    \If{not \DuplicateQuery{$q_i$, $M$}}{
	   \textit{// If $q_i$ has not been processed.}\\
	   $F =$ \IR{$q_i$, $a_i$}; $M.\add(q_i)$\;
	   \lIf{not F == ``$Pass$''}{
	    \textbf{return} $F$
	   }
	   }
	}
    \textbf{return} ``$Finish$'' \;
    }

    \SetKwFunction{Main}{Main}
    \SetKwProg{Fn}{Function}{:}{}
    \Fn{\Main{$Q$,$F$}}{
	\While{not (F == ``Finish'' or $r$ > $r_{max}$)}{
	    $\boldsymbol{CoQ}$ = \ChainGenerate{$Q$, $F$}\;
	    \textit{// LLM generate the new Chain-of-Query $\boldsymbol{CoQ}$.}\\
	    $\boldsymbol{T}$.\AddChild{$\boldsymbol{CoQ}$}; 
	    \textit{// Add the new branch to $\boldsymbol{T}$.}\\
	    $F =$ \Traverse{$\boldsymbol{CoQ}$}; 
	    \textit{// Interact with IR.}\\
	    $r = r+1$; 
	    \textit{// Update the number of interaction rounds $r$.}
	}
    \textbf{return} \Tracing{$\boldsymbol{T}$, $R$} %\tcp*{Generate final result ${G}$.}
    }
\label{alg}

\end{algorithm}
\subsection{Interaction with Information Retrieval} \label{interaction}
In each round of interaction, LLM passes the generated CoQ to IR. IR verifies and completes the information for each node $(q_i, a_i)$ of CoQ and feeds back to LLM to help it generate more correct CoQ as the new branch of ToR (Tree-of-Reasoning). Besides, IR records the corresponding retrieved documents for each node of CoQ as its supporting documents, which enhances the traceability of LLM-generated content. The description of interaction is shown in Algorithm~\ref{alg}. IR interacts with each node $(q_i, a_i)$ of CoQ, retrieves the Top-1 document $d_i$ for $q_i$ as the supporting document, and judges whether to verify or complete it according to the type of $q_i$. When all the queries of CoQ do not need to be corrected or completed, or the maximum number of interaction rounds is reached, the interaction ends. SearChain traces back the correct reasoning path of ToR and refers to each node of the path to generate the final content with marked references to supporting documents for knowledge of each node.

\textbf{Verification. }Verification aims to guarantee the correctness of $a_i$ in each node $(q_i,a_i)$ of CoQ based on the external knowledge base, which improves the accuracy and credibility of generated content. Specifically, given the retrieved Top-1 document $d_i$ for $q_i$, a Reader~\cite{dpr} that has been trained on open-domain QA datasets~\cite{dpr} is used to extract the answer $g$ for $q_i$ from $d_i$ with its confidence $f$ ($f$ is a predicted value that measures whether $g$ can answer $q_i$):
\begin{align}
\centering
    s = \arg\max(\textrm{softmax}(\textbf{H}\textbf{w}_{s})), e = \arg\max(\textrm{softmax}(\textbf{H}\textbf{w}_{e})), 
    \nonumber
\end{align}
\begin{align}
    g=d_i[s:e], f = \textbf{H}_{[CLS]} \textbf{w}_{f}, (\textbf{w}_{s}, \textbf{w}_{t}, \textbf{w}_{f}\in\mathbb{R}^{E}),
    \nonumber
\end{align}
where $\textrm{\textbf{H}}\in\mathbb{R}^{L\times E}$ is the sequence of last hidden states for the input text ``[CLS]$q_i$[SEP]$d_i$'', $L$ is the length and $E$ is hidden dimension. $\textbf{H}_{\textrm{[CLS]}}$ is the last hidden state of [CLS] token. Then, SearChain judges whether the answer $a_i$ given by LLM is consistent with the retrieved information according to (1) whether $g$ appears in $a_i$ (for short-form generation tasks such as multi-hop QA and slot filling) or (2) whether ROUGE~\cite{rouge} between $a_i$ and $d_i$ is greater than the threshold $\alpha$ (for long and free-form generation tasks such as ELI5~\cite{eli5}). If $a_i$ is not consistent with retrieved information and the Reader is confident enough ($f > \theta$, $\theta$ is a threshold to alleviate the negative effect of IR on LLM), a prompt is constructed to help LLM correct the answer $a_i$. The template of the prompt is: \sethlcolor{hl}\hl{\textit{``According to the Reference, the answer for {$q_i$} should be {$g$}, you can change your answer and continue constructing the reasoning chain for [Question]: {$Q$}. Reference: {$d_i$}.''}}. This round is over. LLM receives the feedback of IR, gives the new answer $a'_{i}$ for $q$, and generates a new CoQ with $(q_i,a'_{i})$ as the root node, which is the new branch of ToR.

% \begin{figure}[t]
% \centering
% \includegraphics[width=\linewidth]{prompt2.pdf} % Reduce the figure size so that it is slightly narrower than the column.
% \caption{Prompt design of verification.}
% \label{motivation}
% \end{figure}

\textbf{Completion. }Completion aims to provide LLM with missing knowledge in nodes of CoQ, which improves the accuracy of generated content. Specifically, in CoQ generation (Section~\ref{CoQ}), LLM marks ``[Unsolved Query]'' for the unsolvable query. For the unsolvable query $q_{i}^{*}$, IR extracts the answer $g^{*}$ from retrieved document $d_{i}^{*}$ as described in Verification. Regardless of whether $f$ is greater than the threshold $\theta$, $g^{*}$ and $d_{i}^{*}$ will be fed back to the LLM in the form of a prompt because the LLM cannot solve $q_{i}^{*}$. The template of the prompt is: \sethlcolor{hl}\hl{\textit{``According to the Reference, the answer for {$q_{i}^{*}$} should be {$g^{*}$}, you can give your answer and continue constructing the reasoning chain for [Question]: {$Q$}. Reference: {$d_{i}^{*}$}.''}}. This round is over. LLM receives the feedback, gives the answer $a_i^{*}$ to solve the query $q_{i}^{*}$ and generates a new CoQ with ($q_{i}^{*}$, $a_i^{*}$) as the root node, which is the new branch of ToR.

% \begin{wrapfigure}{r}{0.5\textwidth}
% \centering
% \includegraphics[width=\linewidth]{reference.pdf} % Reduce the figure size so that it is slightly narrower than the column.
% \caption{Final content with reference.}
% \label{refe}
% \end{wrapfigure}

\textbf{Tracing. }Tracing aims to generate the reasoning process and mark references to supporting documents for each reasoning step, which is used as the final content returned to the user. This improves the traceability of each knowledge in the generated content. Specifically, SearChain records the documents retrieved for each node on the correct reasoning path of Tree-of-Reasoning as the supporting documents. SearChain prompts LLM to generate the final content by referring to nodes on the correct path and marking references to the supporting documents for the corresponding sub-fragments of the generated content (final content of Figure~\ref{intro}). The prompt is \sethlcolor{hl}\hl{\textit{``You can try to generate the final answer for the [Question] by referring to the [Query]-[Answer] pairs, starting with [Final Content]. [Query $1$]: {$q_1$}. [Answer $1$]: {$a_1$} ... [Query $m$]: {$q_m$}. [Answer $m$]: {$a_m$}.''}}. This design enables the user to acquire the related documents of the knowledge involved in each step of reasoning. We believe that it is a promising task to mark references to supporting documents on sub-fragments of complex content generated by LLM. Our approach provides a novel and effective approach to solve this task by retrieving supporting documents for each sub-questions involved in the reasoning process of LLM without any supervised data (texts with citation annotations) and training of the LLM.

\textbf{Node-identify Depth-first Search. } Compared with previous retrieval-augmented methods, interaction with IR in SearChain forms a novel reasoning path: node-identify Depth-first Search on a tree. In each generation, LLM generates a CoQ to perform continuous reasoning on complex questions until the final answer is generated or an unsolvable sub-question is encountered. This can be seen as a part of Depth-first Search (DFS). However, different from traditional DFS algorithm~\cite{dfs}, "node-identify" in SearChain means that when a search in one direction is terminated, SearChain does not return to its parent node, but dynamically identifies the node that needs to be corrected or completed via verification and completion in IR and re-generates a new CoQ started with this node. The interaction process between IR and LLM in SearChain is the process of constructing a tree using node-identify DFS, which enables LLM to dynamically modify the reasoning direction.

\section{Experiments}
\label{exp}
% In this section, we compare SearChain with recent related baselines on complex knowledge-intensive tasks and conduct the analysis.

\subsection{Experimental Setup}
\subsubsection{\textbf{Datasets and Evaluation Metric}} We select four classic complex knowledge-intensive tasks including multi-hop question-answering (HotpotQA (HoPo)~\cite{hotpotqa}, Musique (MQ)~\cite{musique}, WikiMultiHopQA (WQA)~\cite{wikimultihop} and StrategyQA (SQA)~\cite{strategyqa}), slot filling (zsRE~\cite{zs}, T-REx~\cite{trex}), fact checking (FEVER~\cite{fever}) and long-form question-answering (ELI5~\cite{eli5}). These tasks require LLM to perform multi-step reasoning on complex questions, and each step requires corresponding knowledge to solve. As for the evaluation metrics, for ELI5 whose ground truth is long and free-form, we use ROUGE-L~\cite{rouge} as the metric. For other tasks, we use whether the ground truth answer is contained within the generated answer (i.e, cover-EM~\cite{cem}) as the metric. Following DSP~\cite{dsp} and Self-Ask~\cite{selfask}, we evaluate the model on full development datasets of MQ and HoPo, BIG-bench~\cite{bigbench} datasets on SQA and subsets of WQA, zsRE, T-REx, FEVER and ELI5 (each subset has 1.2k questions).

% Following DSP~\cite{dsp} and Self-Ask~\cite{selfask}, we evaluate the model on full development datasets of MQ and HoPo, BIG-bench~\cite{bigbench} datasets on SQA, subsets of WQA, zsRE, T-REx, FEVER and ELI5 (each subset has 1.2k questions).

\subsubsection{\textbf{Baselines.}}
Our baselines can be divided into two categories, one is about improving the reasoning ability of LLM on complex tasks (CoT~\cite{chain}, CoT-SC~\cite{chain2}, Auto-CoT~\cite{chain5}, Recite-and-answer~\cite{recitationaugmented} and Least-to-Most~\cite{leasttomost}), and the other is not only introducing IR to LLM but also improving the reasoning ability (Direct\footnote{Retrieve documents and provide them to LLM in a prompt.}, Self-Ask~\cite{selfask}, ToolFormer\footnote{Perform ToolFormer on \textit{gpt-3.5-turbo} via in-context learning.}~\cite{toolformer}, React~\cite{react}, DSP~\cite{dsp}, Verify-and-Edit (combined with CoT-SC)~\cite{vf} and Tree-of-Thought~\cite{tot}). AgentGPT and PS~\cite{ps} use Plan-and-Solve paradigm, we also reproduce this as one of the baselines.
%Details are introduced in Appendix.

\begin{table*}[t]
\centering
\setlength\tabcolsep{11pt}%调列距
\renewcommand\arraystretch{1}
\caption{Performance of SearChain and baselines on complex knowledge-intensive tasks. \textbf{Bold} denotes the best result in different settings. FC: Fact Checking, LFQA: Long-Form QA. Metric for LFQA: ROUGE-L. Metric for others: cover-EM.}
\scalebox{1}{
\begin{tabular}{lllllllll}
\toprule
 &  \multicolumn{4}{c}{Muti-Hop QA} & \multicolumn{2}{c}{Slot Filling} & FC & LFQA \\

                     & HoPo       & MQ        & WQA  & SQA   & zsRE  & T-REx & FEV. & ELI5 \\\hline
\multicolumn{9}{c}{Without Information Retrieval}                                         \\
Direct Prompting      & 31.95          & 5.91           & 25.82          & 66.25  & 22.75 & 43.85  & 73.45 & 21.90        \\
%Zero-shot CoT      & 0 & 31.95          & 5.91           & 25.82          & 66.25          \\
Auto-CoT               & 33.53          & 10.55          & 29.15           & 65.40   & 21.30 & 43.98 & 76.61 & 21.55      \\
CoT                    & 35.04          & 9.46          & 30.41          & 65.83    & 22.36 & 44.51 & 76.98 & 21.79      \\
CoT-SC                 & 36.85          & 10.02          & 32.68          & 70.84   & 24.74 & 46.06 & 77.15 & 22.05        \\
Recite-and-answer     & 36.49          & 10.97          & 32.53          & 70.47 & 24.98 & 46.14 & \textbf{77.35} & 22.10 \\
Self-Ask w/o IR  & 33.95          & 11.10          & 35.65         & 65.45    & 20.16 & 44.71 & 75.31 & 21.73      \\
Least-to-Most           & 34.05          & 11.45          & 32.88          & 65.78    & 21.86 & 44.98 & 75.98 & 21.95      \\
Plan-and-Solve & 36.33 & 12.95 & 35.68 & 73.21 & 25.15 & 47.58 & 77.08 & 22.23\\
SearChain w/o IR     & \textbf{38.36} & \textbf{13.61} & \textbf{40.49} & \textbf{75.62} & \textbf{30.14} & \textbf{52.69} & 77.06 & \textbf{22.54}  \\ \hdashline
\multicolumn{9}{c}{Interaction with Information Retrieval}                                 \\
Direct Retrieval      & 34.09          & 10.22          & 30.01          & 66.78  & 52.29 & 59.28 & 78.25 & 23.40        \\ 
ToolFormer    & 36.75          & 12.98          & 35.49          & 67.02     & 51.35 & 59.17 & 80.79 & 23.05 \\
Self-Ask          & 40.05          & 14.28          & 39.58          & 67.65      & 50.51 & 59.12 & 79.41 & 23.25     \\
Plan-and-Solve w/ IR  & 41.65 & 15.07 & 42.05 & 74.58 & 52.15 & 60.03 & 81.04 & 24.56\\
React → CoT-SC          & 43.15          & 15.49          & 40.36          & 70.43   & 53.27 & 60.42 & 80.59 & 24.05        \\
Verify-and-Edit          & 44.03          & 15.57          & 40.83          & 71.09   & 53.95 & 61.10 & 80.67 & 23.80        \\
Tree-of-Thought w/ IR    & 50.65         & 15.61          & 42.49          & 72.55    & 54.88 & 62.40 & 81.03 & 24.20      \\
DSP                     & 51.97         & 15.83          & 43.52          & 72.41    & 54.35 & 61.32 & 80.65 & 23.46      \\
SearChain                    & \textbf{56.91} & \textbf{17.07} & \textbf{46.27} & \textbf{76.95} & \textbf{57.29} & \textbf{65.07} & \textbf{81.15} & \textbf{25.57} \\
\quad - w/o Verification & 46.11 & 14.70 & 42.67 & 75.98 & 43.58 & 55.46 & 78.79 & 22.98\\
\quad - w/o Completion & 53.05 & 15.86 & 43.64 & 76.53  & 45.78 & 56.03 & 80.03 & 25.02\\
\toprule
\end{tabular}}
\label{main results}
\end{table*}

\subsubsection{\textbf{Implementation}} The large language model we used is \textit{gpt-3.5-turbo} provided from API of OpenAI\footnote{https://openai.com/api/} and the retrieval model we used is ColBERTv2~\cite{colbertv2} (following DSP). IR model infers on one Tesla V100 GPU. For HotpotQA, we use Wikipedia 2017 as the corpus, which is provided by~\cite{hotpotqa} in full-wiki setting. For the other datasets, we use the large-scale passage collection built on Wikipedia as the corpus~\cite{dpr,xu2023berm}. Baselines with information retrieval are in the same setting as SearChain. We reproduce all baselines on \textit{gpt-3.5-turbo} following the settings in their papers. The maximum number of interaction rounds $r_{max}$ is $5$. The thresholds $\alpha$ and $\theta$ are set as 0.35 and 1.5 respectively. As for the selection of confidence threshold ($\theta$), we initialize the initial value of the confidence threshold (1.0) based on prior knowledge and gradually increase the value with a step size of 0.1. We validate the F1-score on the mixed open-domain QA datasets (NQ, TriviaQA, WebQ, and TREC) after each value change. We find that when the confidence threshold is 1.5, the highest F1-score can be achieved so we set the confidence threshold as 1.5. As for the selection of ROUGE threshold ($\alpha$), we determine this value by observing the ROUGE relationship between the generated text and the ground truth in the few examples in in-context learning. Our further experiments show that when the value range of ROUGE threshold is between 0.3 and 0.5, the performance change on ELI5 is not obvious. Details of prompts and experiments are introduced in Section~\ref{app-exp-detail} of Appendix.

\subsection{Main Results}
Performance on knowledge-intensive tasks is shown in Table~\ref{main results}.

\textbf{(1) Effect of Chain-of-Query. }CoQ is the reasoning chain for complex questions in SearChain. We compare it with recent competitive baselines in the setting without IR. SearChain w/o IR outperforms all baselines based on CoT (CoT, Auto-CoT, CoT-SC and Recite-and-answer), which indicates that focusing on constructing a global reasoning chain consisting of sub-questions is better than just giving intermediate reasoning results. SearChain w/o IR outperforms Self-Ask w/o IR and Least-to-Most, which indicates that it is more effective to focus on constructing a global reasoning chain at each inference (global perspective) than generating and answering sub-questions step by step (local perspective). 

\textbf{(2) Effect of interaction with IR. }In the setting with interaction with IR, SearChain again outperforms all the baselines. The paradigm of first generating global CoQ, and then IR interacting with each node of CoQ ensures the coherence of LLM reasoning. This solves the problem in Self-Ask, DSP and React. Besides, SearChain decouples the knowledge of LLM and IR. IR judges whether to provide information to LLM according to the confidence and the flag of the node on CoQ, which effectively alleviates misleading LLM. Last but not least, baselines reason in the one-dimensional chain. They cannot dynamically modify the reasoning direction. Interaction with IR in SearChain transforms the reasoning path from a chain to node-identify Depth-first Search on a tree, which enables LLM to dynamically modify the reasoning direction.

\subsection{Analysis}
\label{analysis}
In this section, we discuss and demonstrate the advantages of SearChain compared to baselines in detail. First, we analyze the source of the knowledge of SearChain in solving complex questions. Second, while we analyze the positive effect of IR on LLM in solving difficult questions, we also demonstrate that our method can better mitigate the negative effect of IR on LLM. Third, we show the advantages of SearChain compared to baselines in terms of reasoning and tracing capabilities. Last but not least, we perform efficiency analysis to show our method significantly
improves task performance with no significant increase in time consumption.

\begin{table}[t]
\centering
\setlength\tabcolsep{7pt}%调列距
\renewcommand\arraystretch{1}
\caption{Distribution of knowledge sources.}
\scalebox{1}{
\begin{tabular}{lllll}
\toprule
Knowledge Src.   & HoPo & MQ & WQA  & SQA \\ \hline
LLM    & 74.56\%    & 78.83\%   & 75.83\% & 94.98\%    \\
Corrected by IR  & 20.94\%    & 14.60\%    & 18.96\% & 2.78\%    \\
Completed by IR & 4.50\%    & 6.57\%   & 5.21\% & 2.24\%    \\ \toprule
\end{tabular}
}
\label{source}
\end{table}

\begin{table}[t]
  \centering
\caption{Positive and negative effects of IR on LLM.}
  \subtable[Accuracy on $\mathbb{S}_{IR}$ and $\mathbb{S}$ (positive effect $\uparrow$).]{
    \centering
    \setlength\tabcolsep{10.5pt}%调列距
    \renewcommand\arraystretch{1.05}
    \scalebox{1}{
    \begin{tabular}{lllll}
    \toprule
                      & HoPo & MQ & WQA  & SQA \\ \hline
    w/o IR ($\mathbb{S}$)      & 38.36    & 13.61   & 40.49 & 75.62    \\
    w/o IR ($\mathbb{S}_{IR}$) & 31.38    & 10.20    & 32.60 & 68.96    \\
    w IR ($\mathbb{S}_{IR}$)   & 60.86    & 18.49   & 50.52 & 78.42    \\ \toprule
    \end{tabular}
    }
    %\subcaption{(a) Accuracy on $\mathbb{S}_{IR}$ and $\mathbb{S}$.}
    \label{accuracy}
  }
  \subtable[Percentage that IR misleads LLM (negative effect $\downarrow$).]{
    \centering
    \setlength\tabcolsep{6.5pt}%调列距
    \renewcommand\arraystretch{1}
    \scalebox{1}{
    \begin{tabular}{lllll}
    \toprule
             & HoPo & MQ & WQA & SQA \\ \hline
    Self-Ask & 15.76    & 14.32   & 25.76          & 10.29      \\
    React & 17.68 & 15.22 & 25.99     & 10.03  \\
    Plan-and-Solve w/ IR & 16.42   & 15.25   & 22.31 & 7.59 \\
    Verify-and-Edit & 9.78    & 10.75   & 16.44 & 6.52 \\
    Tree-of-Thought w/ IR & 12.07    & 13.25   & 20.52 & 8.46 \\
    DSP      & 14.72    & 14.03   & 24.31         & 9.22 \\
    SearChain     & 6.33     & 6.50    & 12.71          & 5.31       \\ \toprule
    \end{tabular}
    }
    %\subcaption{(b) Percentage that IR misleads LLM.}
    \label{mislead}
   }
  \label{pos_and_neg}
\end{table}

\subsubsection{\textbf{Knowledge Decoupling}} We analyze the knowledge sources on the four multi-hop QA datasets. Specifically, we classify knowledge sources into three categories: (1) knowledge of LLM, (2) knowledge that corrected by IR in verification, and (3) knowledge that LLM does not know and is provided by IR in completion. We use node of ToR as the statistical granularity to calculate the percentage of nodes from these three sources in the total nodes respectively. The experimental results are shown in Table~\ref{source}. It is worth noting that even though most of the knowledge comes from LLM, this knowledge is also verified by IR. IR only corrects the inconsistent answer given by LLM when it is confident enough and provides LLM with the missing knowledge, which alleviates the negative effect of IR on LLM and improves the utilization of retrieved information. On StrategyQA, LLM has memorized most knowledge that IR can retrieve, so IR provides less knowledge than other datasets. 

\subsubsection{\textbf{Positive and Negative Effects of IR on LLM}} 
\textbf{(1) Positive.} In SearChain, IR can identify the trouble of LLM and effectively help LLM to correct the answers and acquire missing knowledge. We select the questions ($\mathbb{S}_{IR}$) that IR helps to correct or provide knowledge from the datasets used in Table~\ref{main results} ($\mathbb{S}$) and evaluate the accuracy of SearChain on $\mathbb{S}_{IR}$. We also evaluate the accuracy of SearChain w/o IR on $\mathbb{S}_{IR}$. The results in Table~\ref{pos_and_neg}(a) show that w/o IR performs worse on $\mathbb{S}_{IR}$ than on $\mathbb{S}$, which indicates that LLM does have trouble with the questions that require IR help. w/ IR performs better on $\mathbb{S}_{IR}$, which indicates that IR effectively identifies and solves the trouble of LLM. \textbf{(2) Negative.} Section~\ref{intro} points out the risk of IR misleading LLM when there is a conflict in the knowledge of IR and LLM. We select the questions ($\mathbb{S}_t$) that LLM can give the correct answers to and count the percentage that LLM gives incorrect answers after adding IR on $\mathbb{S}_t$. Table~\ref{pos_and_neg}(b) shows SearChain effectively mitigates the negative effect of IR on LLM. SearChain uses the confidence of IR and the information of CoQ to judge whether to correct LLM or provide LLM with its missing knowledge.

\begin{table}[t]
\setlength\tabcolsep{7pt}%调列距
\renewcommand\arraystretch{1}
\caption{Number of reasoning steps. SearChain tries more for unsolvable sub-questions to achieve better accuracy.}
\scalebox{1}{
\begin{tabular}{lllll}
\toprule
                & 2-hop & 3-hop & 4-hop & Accuracy \\ \hline
CoT             & 2.25  & 2.23  & 2.19 & 9.46 \\
Self-Ask w/o IR & 2.04  & 2.21  & 2.15 & 11.10\\
Least-to-Most   & 2.52  & 2.68  & 2.70 & 11.45\\
SearChain w/o IR     & 4.16  & 4.66  & 5.06 & 13.61\\ \toprule
\end{tabular}
}
\label{steps}
\end{table}

\begin{figure}[t]
\centering
\includegraphics[width=1.0\linewidth]{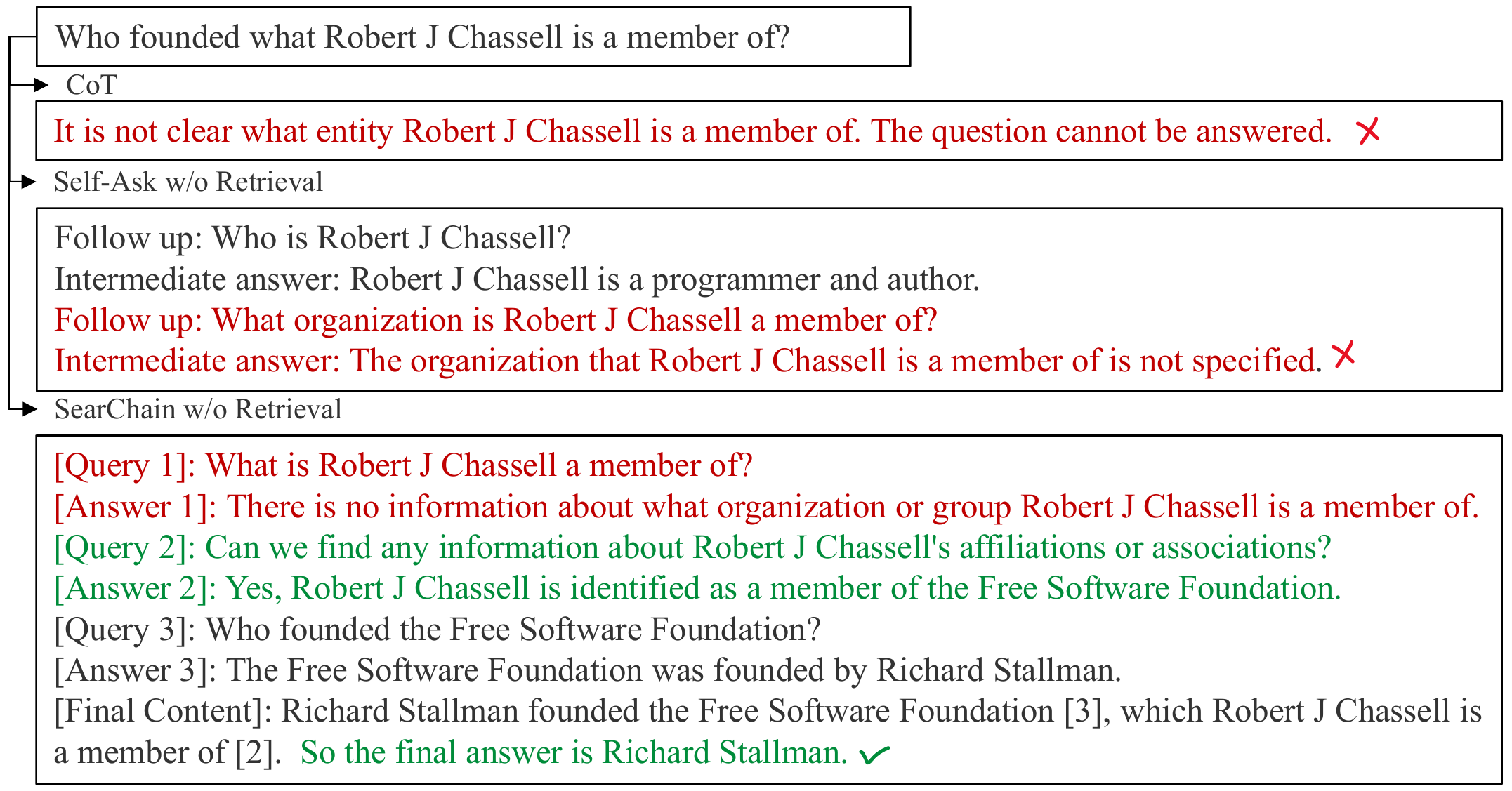} % Reduce the figure size so that it is slightly narrower than the column.
\caption{Case study of the difference between SearChain and baselines for unsolvable sub-questions.}
\label{case}
\end{figure}

\begin{figure*}[t]
\centering
\includegraphics[width=1\linewidth]{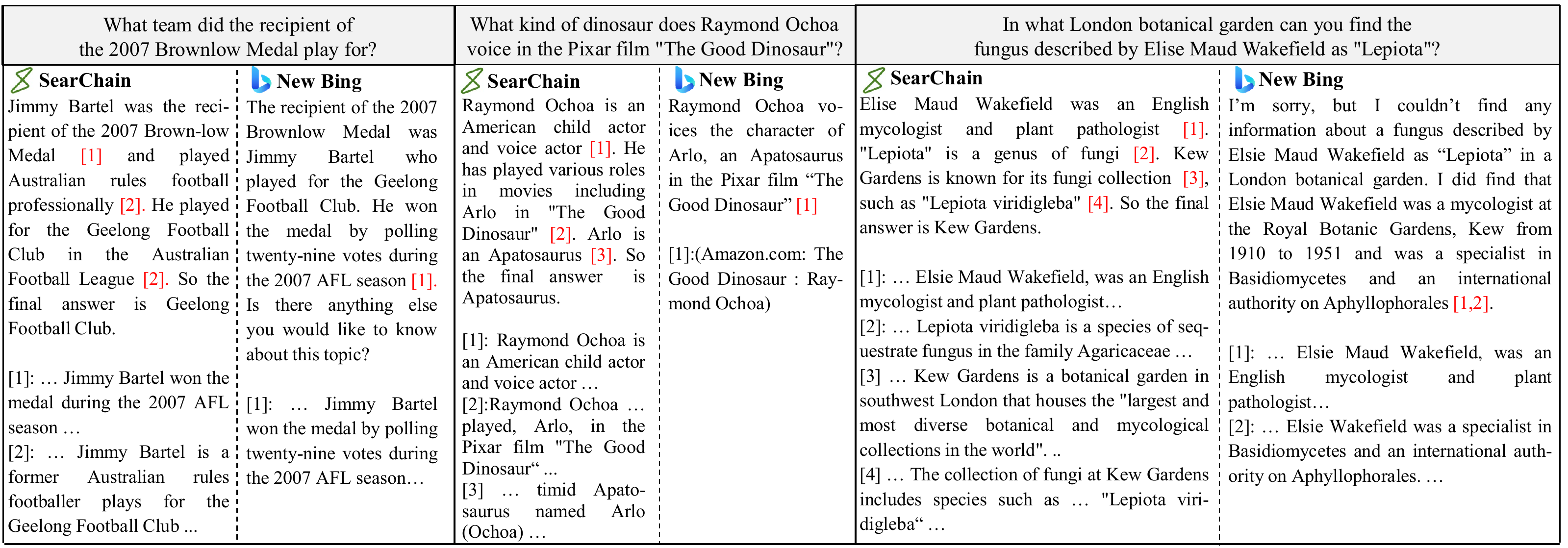} % Reduce the figure size so that it is slightly narrower than the column.
\caption{Case study of SearChain and New Bing in marking references to supporting documents.}
\label{vsnewbing}
\end{figure*}

\subsubsection{\textbf{CoQ vs Baselines in Reasoning}} \label{reasoning}
CoQ performs better on reasoning complex questions than the baselines. In addition to Table~\ref{main results}, we further analyze the reasoning ability from two aspects: 

\textbf{(1) Number of Reasoning Steps.} We analyze the number of reasoning steps in different methods in the setting without IR. We conduct the experiment on Musique because Musique has more complex questions. Table~\ref{steps} shows the average number of reasoning steps on questions with different hops. Our method has more reasoning steps, and the number of reasoning steps increases with the hops of the question. This shows that our method has a better perception of the complexity of the questions.

\textbf{(2) Solving Difficult Sub-questions.} The baselines focus on solving local sub-questions while ignoring the global planning of the reasoning chain. This leads LLM to tend to stop reasoning rather than try more when a sub-question cannot be solved. In our method, LLM acts as a commander that plans a global reasoning chain that can solve the complex question, when a sub-question cannot be solved, even without the help of IR, LLM can try to further decompose or rewrite the sub-question to continue reasoning. It is because our method focuses on building a global chain that can solve the complex question (global perspective), rather than answering or generating the sub-questions step by step (local perspective). Case study in Figure~\ref{case} shows that CoT and Self-Ask stop the reasoning while SearChain continues reasoning by rewriting the sub-question. More reasoning steps in Table~\ref{steps} also support that SearChain can try more for difficult sub-questions. More case studies are shown in Section~\ref{app-reason} of Appendix.

\subsubsection{\textbf{SearChain vs New Bing in Tracing}} We compare the performance of SearChain and New Bing in marking references for generated content via case study (Figure~\ref{vsnewbing}). We further propose two metrics to evaluate the Scope of Knowledge Coverage and Accuracy of Marking Position to show traceability more intuitively:
% \begin{itemize}
% \item{} Scope of Knowledge Coverage (SKC) [0, +]: The number of knowledge items marked with supporting documents in the generated content. (statistics)
% \item{} Accuracy of Marking Position (AMP) [0, 1]: The accuracy of the position of the reference mark. That is, whether the references are correctly marked on the sub-fragments for the corresponding knowledge in the generated content. (human evaluation)
% \end{itemize}

\noindent \textbf{$\mathbf{\bullet}$ Scope of Knowledge Coverage (SKC) [0, +]}: The number of knowledge items marked with supporting documents in the generated content. (statistics, SearChain (2.882) is better than New Bing (1.143))

\noindent \textbf{$\mathbf{\bullet}$ Accuracy of Marking Position (AMP) [0, 1]}: The accuracy of the position of the reference marks. That is, whether the references are correctly marked on the sub-fragments for the corresponding knowledge in the generated content. (human evaluation, SearChain (0.80) is better than New Bing (0.45))

\noindent We introduce three humans with master's degrees to participate in our human evaluation and the results show that SearChain can mark references for each knowledge involved in the reasoning process (i.e., correct nodes of CoQ) in a fine-grained manner. While the references given by New Bing do not cover all of the knowledge and cannot be marked on the correct position. More case studies are shown in Section~\ref{app-trac} of Appendix. 

% \begin{table}[t]
% \centering
% \setlength\tabcolsep{15pt}%调列距
% \renewcommand\arraystretch{1}
% \caption{Evaluation of traceability.}
% \scalebox{1}{
% \begin{tabular}{lll}
% \toprule
%      & SKC  & AMP  \\ \hline
% New Bing & 1.143  &  0.45   \\ 
% SearChain & \textbf{2.882} & \textbf{0.80} \\
% \toprule
% \end{tabular}
% }
% \label{evtr}
% \end{table}

\begin{table}[t]
\centering
\setlength\tabcolsep{2pt}%调列距
\renewcommand\arraystretch{1.1}
\caption{Efficiency analysis.}
\scalebox{1}{
\begin{tabular}{lllllc}
\toprule
         & \#$n\downarrow$  & \#$m\downarrow$  & $\#r\downarrow$ & $t (s)\downarrow$ & Perf. (Avg) $\uparrow$ \\ \hline
Self-Ask &  401   & 63   &  2.19  & 6.63 & 46.73      \\
Plan-and-Solve w/ IR &  450   & 71   &  1  & 6.05 & 48.89      \\
React → CoT-SC &  938   & 110   &  2.35  & 8.25 & 48.47      \\
Verify-and-Edit   &  565   & 307   &  2.40  & 13.90 & 48.88 \\
Tree-of-Thought  w/ IR &  622   & 341   &  2.29  & 13.28 & 50.47 \\
DSP      & 1759   & 155   &  2.15   & 10.47 & 50.44    \\
SearChain     & 390  & 189    &  2.21 & 8.52 & 53.29   \\ \toprule
\end{tabular}
}
\label{efficiency}
\end{table}
\subsubsection{\textbf{Efficiency Analysis}} We analyze the running efficiency between SearChain and baselines on the number of words in the input ($n$) and output ($m$) text of LLM, number of rounds of interaction between LLM and IR ($r$) and overall running time ($t$). Table~\ref{efficiency} shows our method significantly improves task performance with no significant increase in time consumption. Most baselines also require multiple rounds of interaction between IR and LLM.

\section{Conclusion}
In this paper, we point out the challenges of introducing IR into LLM from the perspectives of reasoning and knowledge. We then propose a novel framework named SearChain to enable IR and LLM to interact with each other effectively. SearChain not only stimulates the knowledge-reasoning ability of LLM but also uses IR to provide the knowledge that LLM really needs based on the external knowledge base, which improves both accuracy and credibility. Besides, SearChain can mark references to supporting documents for the knowledge involved in the generated content, which improves the traceability of the content. In addition, the interaction between IR and LLM in SearChain transforms the reasoning path from a chain to node-identify Depth-first Search
on a tree, which enables LLM to dynamically modify the reasoning direction. Experimental results on complex knowledge-intensive tasks show that SearChain performs better than all baselines. 

\begin{acks}
This work was supported by the National Key R\&D Program of China (2022YFB3103700, 2022YFB3103704), the National
Natural Science Foundation of China (NSFC) under Grants No. 62276248 and U21B2046, and the Youth Innovation Promotion Association
CAS under Grants No. 2023111.
\end{acks}

\normalem

\bibliographystyle{ACM-Reference-Format}
%% \bibliography{sample-base}
\bibliography{my}
%%
%% If your work has an appendix, this is the place to put it.
\appendix

\section{Appendix}

\subsection{Case Study}
In this section, we compare the performance of SearChain and New Bing~\footnote{https://www.bing.com/new} in adding references to supporting documents for generated content via case study. We also use case study to further analyze why CoQ has stronger reasoning ability than Baselines.

\subsubsection{Case Study for CoQ vs Baselines in Reasoning} \label{app-reason}
\begin{figure}[h]
\centering
\includegraphics[width=\linewidth]{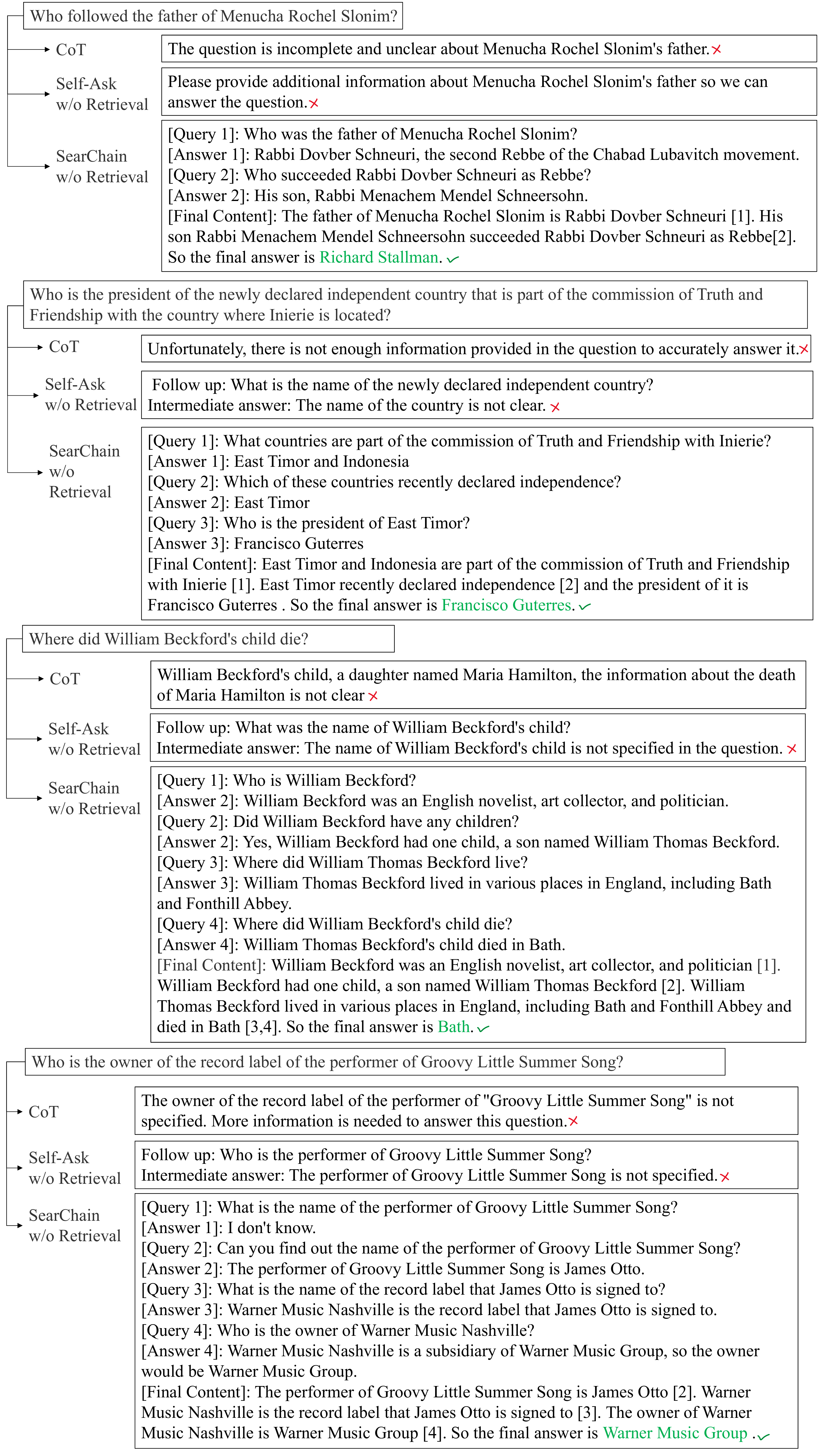}
\caption{Case study for CoQ vs Baselines in Reasoning.}
\label{reasoning2}
\end{figure}

Baselines focus on solving local sub-questions while ignoring the global planning of the reasoning chain, which leads LLM to tend to stop reasoning rather than try more when a sub-question cannot be solved. In our method, LLM acts as a commander that plans a global reasoning chain that can solve the complex question, when a sub-question cannot be solved, even without the help of IR, LLM can try to further decompose or rewrite the sub-question to continue reasoning. It is because our method focuses on building a global chain that can solve the complex question (global perspective), rather than answering or generating sub-questions step by step (local perspective). This makes LLM try more when faced with intermediate difficulties to finally solve complex questions. Case study shown in Figure~\ref{reasoning2} indicates that SearChain continues reasoning but baselines stop.

\subsubsection{Case Study for SearChain vs New Bing in Tracing} \label{app-trac}
\begin{figure}[t]
\centering
\includegraphics[width=1\linewidth]{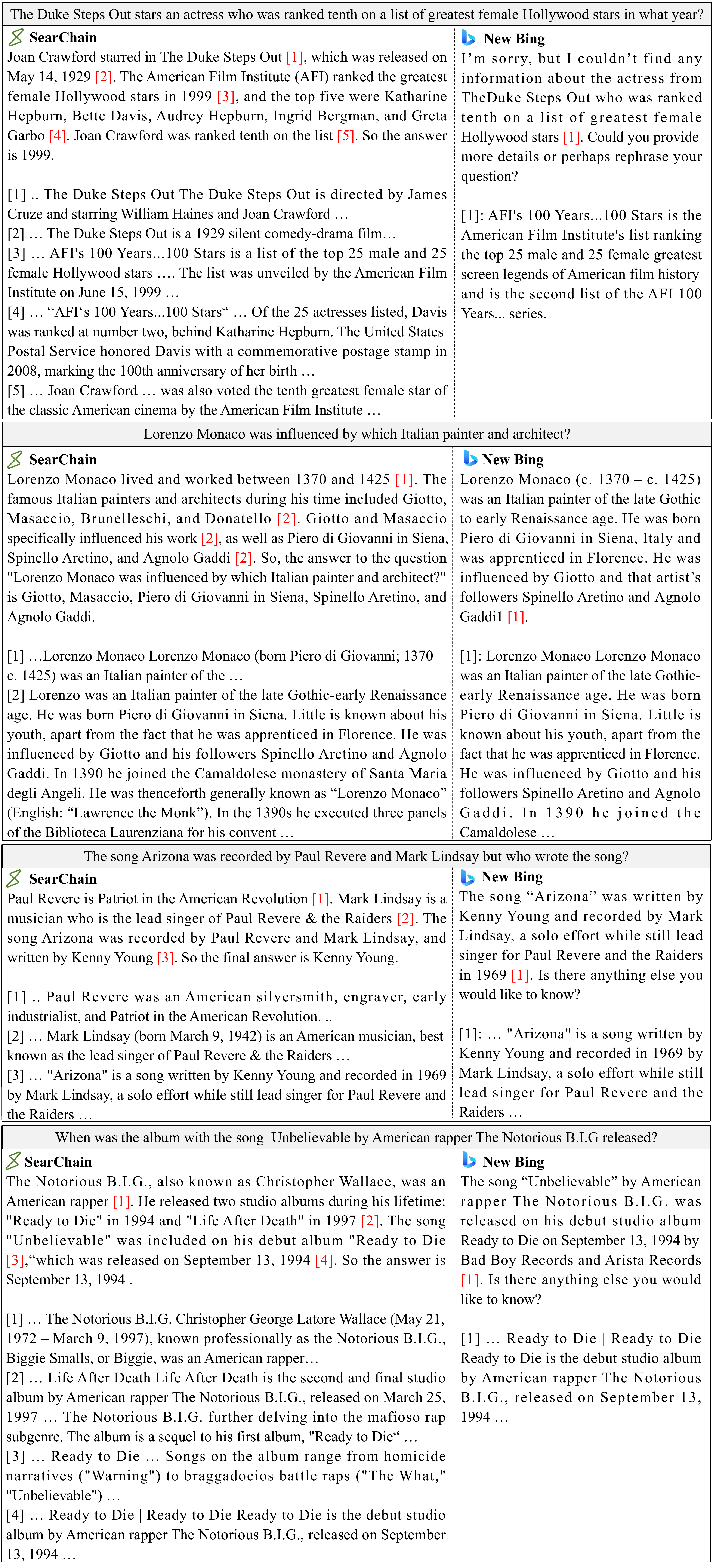} 
\caption{Case study for SearChain vs New Bing in Tracing.}
\label{case6789tracing}
\end{figure}

We compare the performance of SearChain and New Bing in marking references for generated content via case study (Figure~\ref{case6789tracing}). SearChain can mark references for each knowledge involved in the reasoning process (i.e., each correct node of CoQ) in a more fine-grained manner. While references given by New Bing do not cover all of the knowledge, and in some cases New Bing cannot find the knowledge. SearChain provides a novel perspective that decomposes complex multi-step knowledge-intensive tasks into multiple single-step knowledge reasoning problems, retrieving the supporting documents of knowledge for each step of reasoning, and organizing these reasoning steps with their reference marks as final generated content. This enables the supporting documents to cover every knowledge involved in the generated content, which enhances the traceability of the generated content.

\begin{table*}[t]
\centering
\setlength\tabcolsep{8.5pt}%调列距
\renewcommand\arraystretch{1.05}
\caption{Performance of SearChain and DSP on complex knowledge-intensive tasks on Vicuna-13B. \textbf{Bold} denotes the best result in different settings. FC: Fact Checking, LFQA: Long-Form QA. Metric for LFQA: ROUGE-L. Metric for others: cover-EM.}
\scalebox{0.85}{
\begin{tabular}{lllllllll}
\toprule
 &  \multicolumn{4}{c}{Muti-Hop QA} & \multicolumn{2}{c}{Slot Filling} & FC & LFQA \\

                     & HoPo       & MQ        & WQA  & SQA   & zsRE  & T-REx & FEV. & ELI5 \\\hline
\multicolumn{9}{c}{Interaction with Information Retrieval}                                 \\
DSP                     & 25.45         & 9.06          & 27.50          & 62.01    & 33.71 & 49.08 & 73.05 & 22.58      \\
SearChain               & \textbf{29.77} & \textbf{10.59} & \textbf{32.32} & \textbf{63.75} & \textbf{36.86} & \textbf{52.75} & \textbf{75.47} & \textbf{24.05} \\
\toprule
\end{tabular}
}
\label{vicuna}
\end{table*}
\subsection{Performance on Vicuna-13B}
In this section, we compare SearChain with the competitive baseline DSP on Vicuna-13B~\footnote{https://lmsys.org/blog/2023-03-30-vicuna/}, a strong open source large model~\footnote{https://huggingface.co/lmsys/vicuna-13b-delta-v1.1/tree/main} trained by Stanford. The experimental results in Table~\ref{vicuna} show that SearChain again outperforms DSP on Vicuna-13B.

\subsection{Experimental Details} \label{app-exp-detail}
\subsubsection{Threshold Selection}
\begin{table}[t]
\centering
\renewcommand\arraystretch{1.2}
  \caption{Performance change with ROUGE threshold.}
\scalebox{0.85}{
\begin{tabular}{llllll}
\toprule
             & $\alpha=0.30$ & $\alpha=0.35$ & $\alpha=0.40$ & $\alpha=0.45$ & $\alpha=0.50$\\ \hline
Performance   & 25.50 & 25.57 & 25.58 & 25.57 & 25.55 \\ \toprule
\end{tabular}
}
  \label{thre}
\end{table}
As for the confidence threshold ($\theta$), we initialize the initial value of the confidence threshold (1.0) based on prior knowledge and gradually increase the value with a step size of 0.1. We validate the F1-score (a comprehensive metric of the Recall and Precision of judging whether the passage can answer the question) on the mixed open-domain QA datasets (NQ, TriviaQA, WebQ, and TREC) after each value change. We find that when the confidence threshold is 1.5, the highest F1-score can be achieved so we set the confidence threshold as 1.5. As for the ROUGE threshold ($\alpha$), we determine this value by manually observing the ROUGE relationship between the generated text and the ground truth in the few examples in in-context learning. Our further experiments in Table~\ref{thre} show that when the value range of ROUGE threshold is between 0.3 and 0.5, the performance change on ELI5 is not obvious.

\subsubsection{Number of Examples in Prompt}
We show the number of examples in prompt used for in-context learning on different datasets (Table~\ref{shot}). Our method (SearChain) achieves the best performance with fewer learning examples than competitive baselines.

\begin{table}[t]
\centering
\setlength\tabcolsep{2pt}%调列距
\renewcommand\arraystretch{1.1}
\caption{Number of examples in prompt used for in-content learning on different datasets.}
\scalebox{0.85}{
\begin{tabular}{lllllllll}
\toprule
 &  \multicolumn{4}{c}{Muti-Hop QA} & \multicolumn{2}{c}{Slot Filling} & FC & LFQA \\

                     & HoPo       & MQ        & WQA  & SQA   & zsRE  & T-REx & FEV. & ELI5 \\\hline
\multicolumn{9}{c}{Without Information Retrieval}                                         \\
Direct Prompting      & 0          & 0          & 0          & 0  & 0 & 0  & 0 & 0        \\
%Zero-shot CoT      & 0 & 31.95          & 5.91           & 25.82          & 66.25          \\
Auto-CoT              & 4          & 4    & 4       & 6          & 4    & 4 & 4 & 2     \\
CoT                    & 4          & 4    & 4       & 6          & 4    & 4 & 4 & 2      \\
CoT-SC                 & 4          & 4    & 4       & 6          & 4    & 4 & 4 & 2        \\
Recite-and-answer      & 4          & 4    & 4       & 6          & 4    & 4 & 4 & 2 \\
Self-Ask w/o IR        & 4          & 4    & 4       & 6          & 4    & 4 & 4 & 2      \\
Least-to-Most          & 4          & 4    & 4       & 6          & 4    & 4  & 4 & 2      \\
Plan-and-Solve         & 4          & 4    & 4       & 6          & 4    & 4 &4 & 2  \\
SearChain w/o IR       & 2          & 2    & 2       & 6          & 2    & 2 & 4 & 2  \\ \hdashline
\multicolumn{9}{c}{Interaction with Information Retrieval}                                 \\
Direct Retrieval      & 0          & 0          & 0          & 0  & 0 & 0  & 0 & 0        \\ 
ToolFormer    & 4          & 4          & 4          & 6     & 4 & 4 & 4 & 2 \\
Self-Ask          & 4          & 4          & 4          & 6 & 4 & 4 & 4 &2     \\
Plan-and-Solve w/ IR  & 4 & 4 & 4 & 6 & 4 & 4 & 4 & 2\\
React → CoT-SC          & 6          & 4          & 4          & 6   & 4 & 4 & 4 & 2        \\
Verify-and-Edit     & 2 & 2 & 2 & 2 & 2 & 2 & 4 & 2        \\
Tree-of-Thought w/ IR     & 4 & 4 & 4 & 6 & 4 & 4 & 4 & 2        \\
DSP                     & 16         & 8          & 8          & 8    & 8 & 8 & 8 & 2      \\
SearChain            & 2 & 2 & 2 & 2 & 2 & 2 & 4 & 2 \\
\toprule
\end{tabular}
}
\label{shot}
\end{table}

\subsubsection{Prompts in Experiment}
\begin{figure}[h]
\centering
\includegraphics[width=\linewidth]{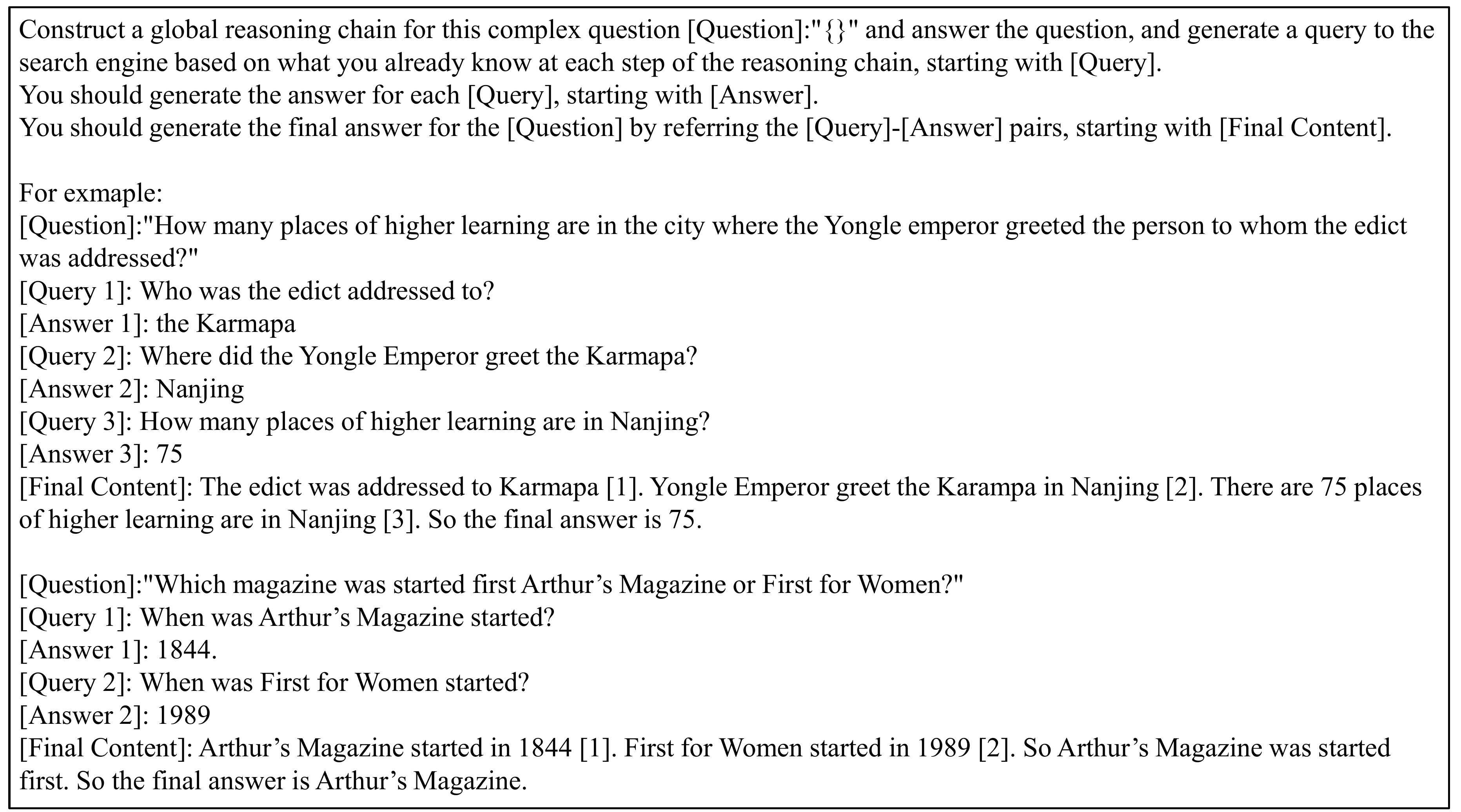}
\caption{Prompt for generating Chain-of-Query on HotpotQA, Musique, WikiMultiHopQA, zsRE and T-REx (in the setting without information retrieval).}
\label{prompt1}
\end{figure}
\begin{figure}[t]
\centering
\includegraphics[width=\linewidth]{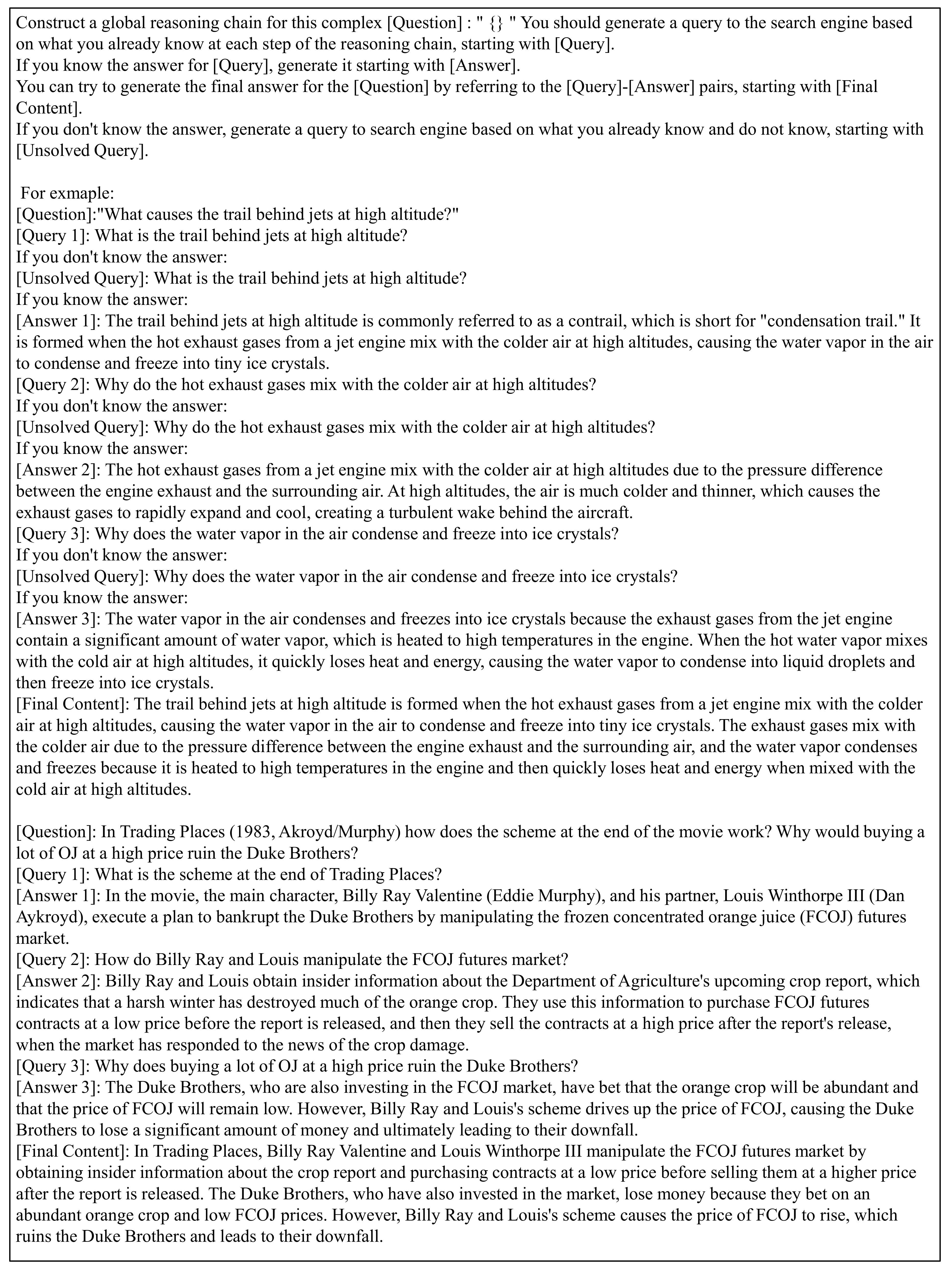}
\caption{Prompt for generating Chain-of-Query at the first round on ELI5 (in the setting with information retrieval).}
\label{prompt7}
\end{figure}

We show the prompt used in experiment on different datasets in Figure~\ref{prompt1} $\sim$ \ref{prompt7}.

\begin{figure}[t]
\centering
\includegraphics[width=\linewidth]{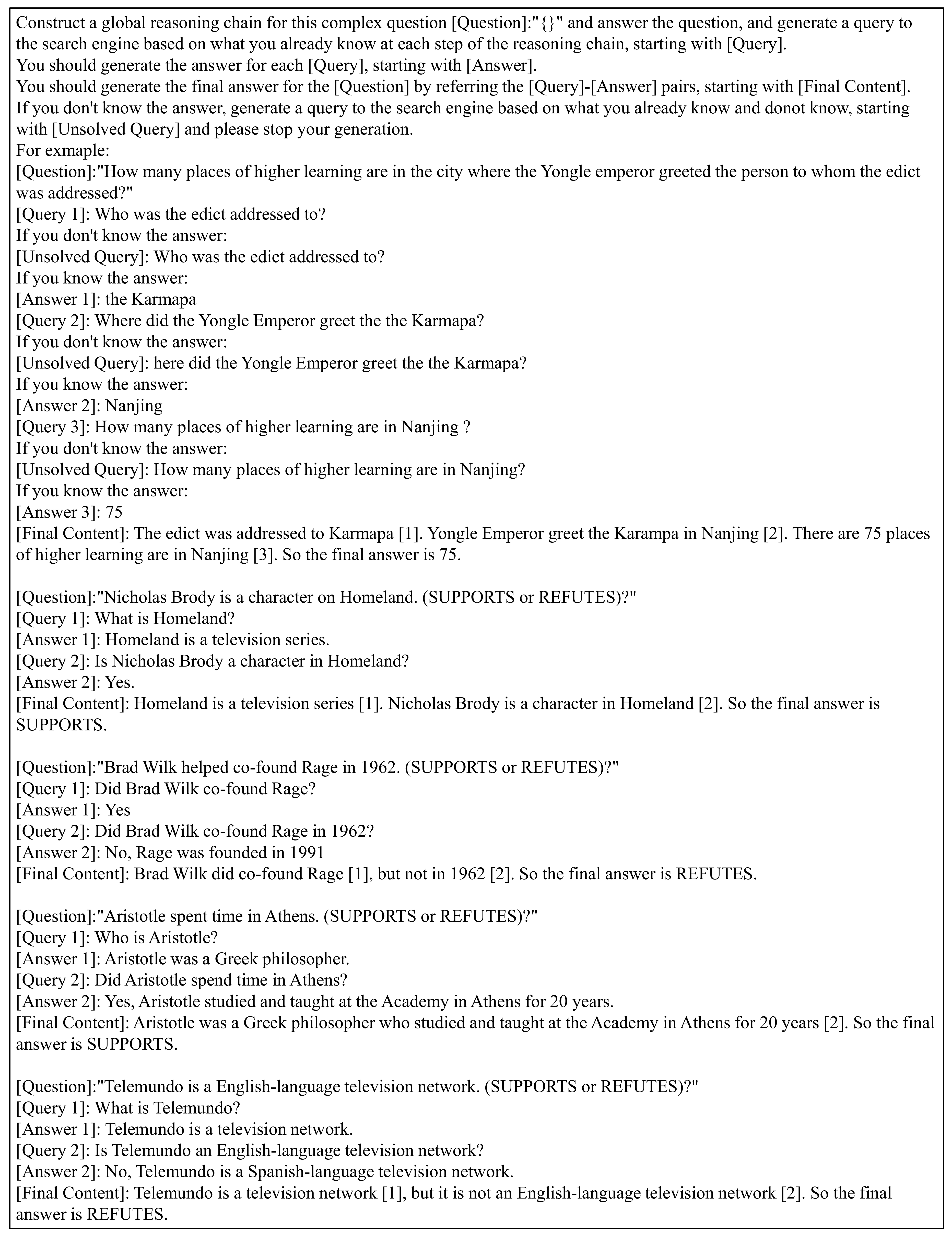}
\caption{Prompt for generating Chain-of-Query at the first round on FEVER (in the setting with information retrieval).}
\label{prompt8}
\end{figure}

\begin{figure}[t]
\centering
\includegraphics[width=\linewidth]{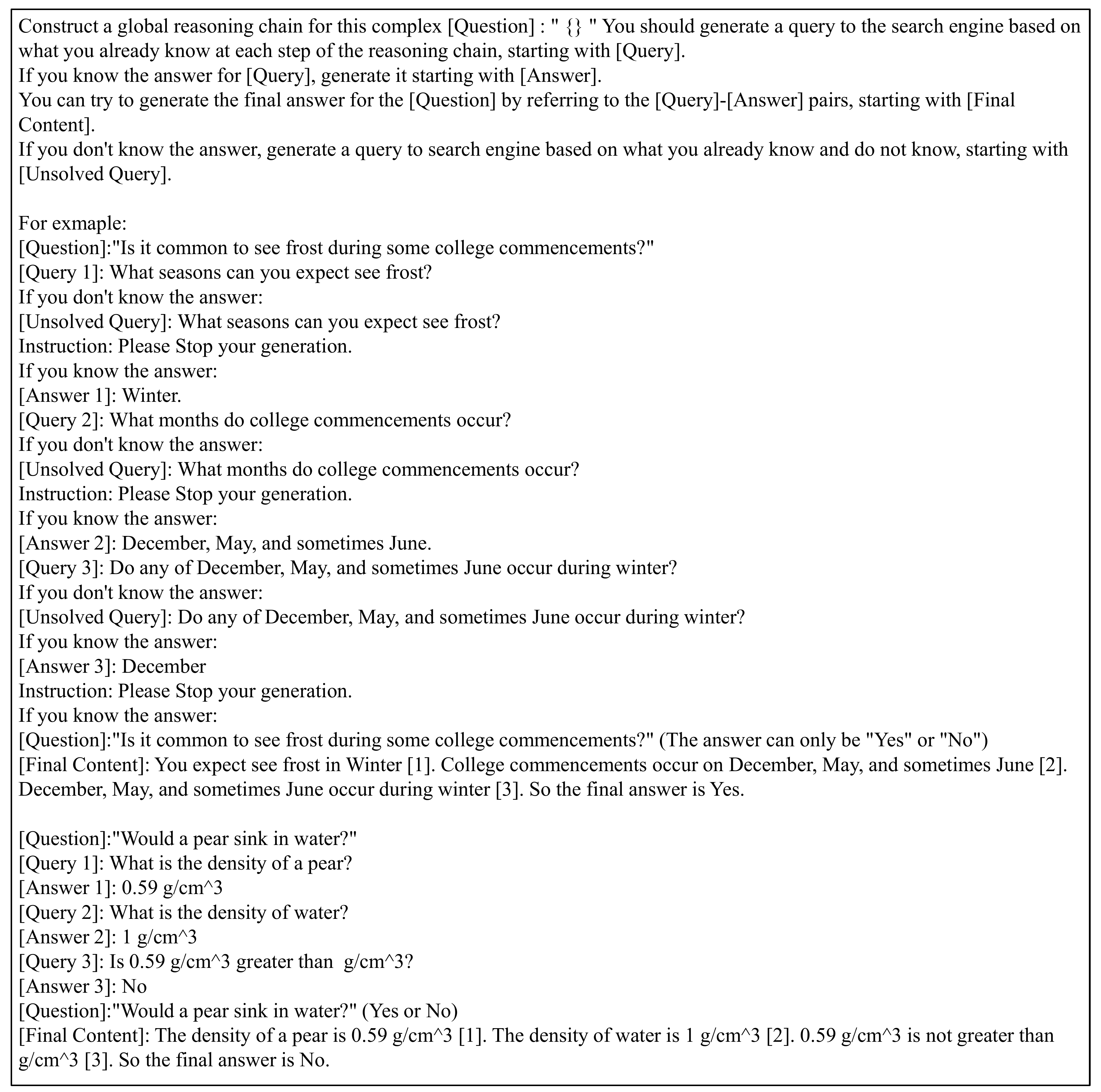}
\caption{Prompt for generating Chain-of-Query at the first round on StragegyQA (in the setting with information retrieval).}
\label{prompt6}
\end{figure}

\begin{figure}[t]
\centering
\includegraphics[width=\linewidth]{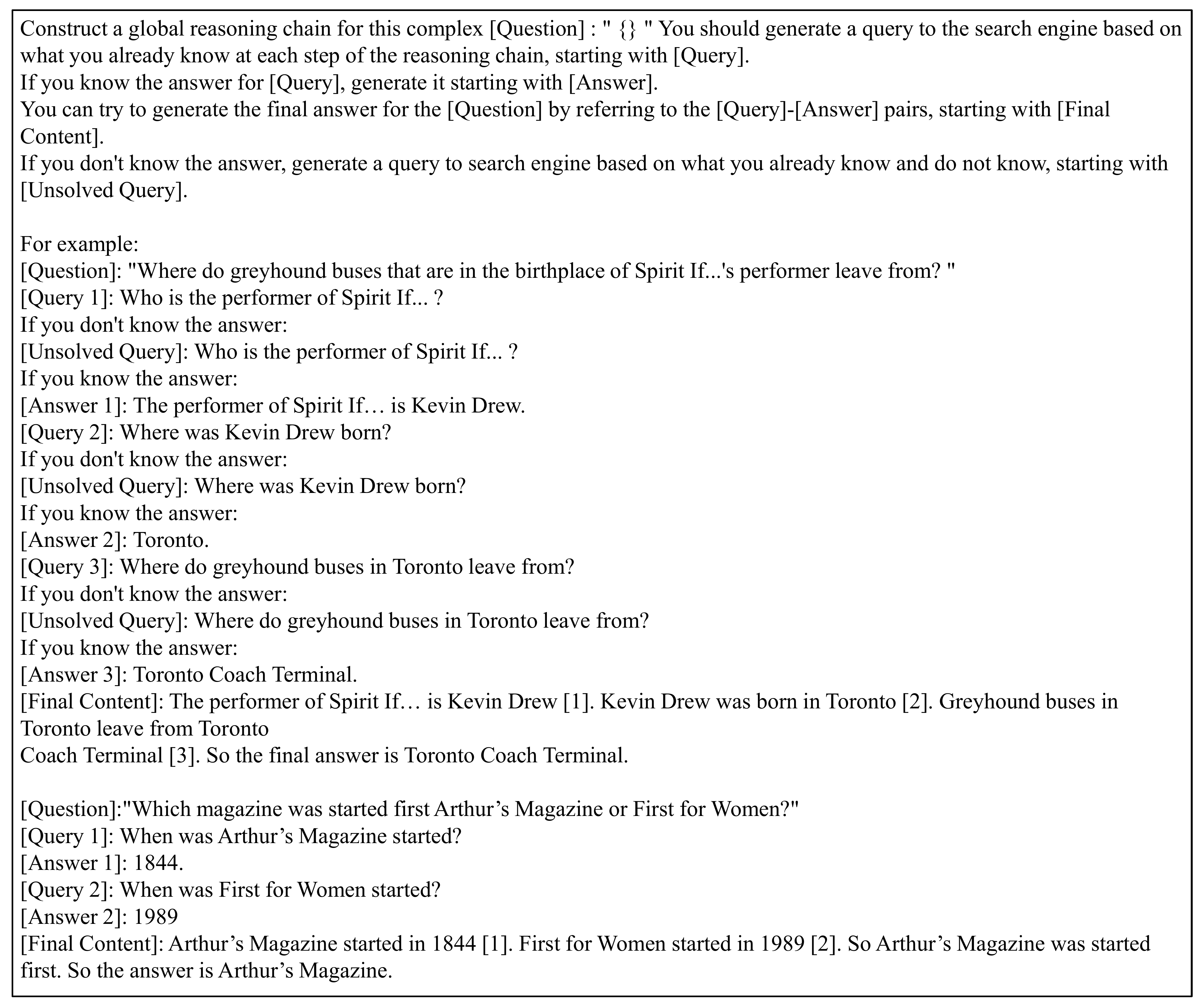}
\caption{Prompt for generating Chain-of-Query at the first round on HotpotQA, Musique, WikiMultiHopQA, zsRE and T-REx (in the setting with information retrieval).}
\label{prompt5}
\end{figure}

\end{document}